# Interpolatron: Interpolation or Extrapolation Schemes to Accelerate Optimization for Deep Neural Networks

**Guangzeng Xie**                                           SMSXGZ@PKU.EDU.CN
*School of Mathematical Sciences*
*Peking University*
*Beijing 100871, China*

**Yitan Wang**                                              YITANWANG@PKU.EDU.CN
*School of Electronic Engineering and Computer Science*
*Peking University*
*Beijing, PEK 100871, China*

**Shuchang Zhou**                                           ZSC@MEGVII.COM
*Megvii*
*Beijing, PEK 100871, China*

**Zhihua Zhang**                                            ZHZHANG@MATH.PKU.EDU.CN
*School of Mathematical Sciences*
*Peking University*
*Beijing 100086, China*

## Abstract

In this paper we explore acceleration techniques for large scale nonconvex optimization problems with special focuses on deep neural networks. The extrapolation scheme is a classical approach for accelerating stochastic gradient descent for convex optimization, but it does not work well for nonconvex optimization typically. Alternatively, we propose an interpolation scheme to accelerate nonconvex optimization and call the method Interpolatron. We explain motivation behind Interpolatron and conduct a thorough empirical analysis. Empirical results on DNNs of great depths (e.g., 98-layer ResNet and 200-layer ResNet) on CIFAR-10 and ImageNet show that Interpolatron can converge much faster than the state-of-the-art methods such as the SGD with momentum and Adam. Furthermore, Anderson's acceleration, in which mixing coefficients are computed by least-squares estimation, can also be used to improve the performance. Both Interpolatron and Anderson's acceleration are easy to implement and tune. We also show that Interpolatron has linear convergence rate under certain regularity assumptions.

## 1. Introduction

Many machine learning models can be formulated as optimization problems. In general, the problem is solved under a regularized empirical risk minimization (ERM) framework:

$$\min_{\mathbf{x}\in\mathbb{R}^d} f(\mathbf{x}) \triangleq \sum_{i=1}^{n} f_i(\mathbf{x}) + \lambda\phi(\mathbf{x}),$$

where $f_i$ corresponds to the loss function with respect to (w.r.t.) the $i$th training sample, $\phi(\mathbf{x})$ is a penalty function, and $\lambda \geq 0$ is a hyperparamter. Typically, $\phi(\mathbf{x}) \triangleq \|\mathbf{x}\|^2 = \sum_{i=1}^{d} x_i^2$ is employed (Krogh and Hertz, 1992) and it plays a role of weigh decay in the training.

Recently, nonconvex optimization has received much attention due to great developments and successful applications of deep neural networks (DNNs) in which a deep structure makes the objective function $f$ nonconvex (Hinton and Salakhutdinov, 2006; LeCun et al., 2016; Goodfellow et al., 2016). In the scenario of DNNs, moreover, both the number of training samples $n$ and the dimension $d$ are very large. Thus, first-order gradient methods are especially efficient due to their lower time complexity and space complexity. However, computing the full gradients of the objective function on the whole data is still too time-consuming and device-demanding. Alternatively, the stochastic gradient descent (SGD) (Murata, 1999) and its variants are used in practical applications.

Although it is well proved that SGD works well on a variety of deep learning models (Krizhevsky et al., 2012; Hinton et al., 2012; Graves et al., 2013), it suffers from the slow convergence rate. Thus, many variants of SGD, such as the momentum (also called the heavy ball), Nesterov's accelerated gradient (Sutskever et al., 2013), and Adam (Kingma and Ba, 2014), have been successively employed for acceleration. However, these variants also bring some dissatisfactory issues, such as numerical instability and more hyperparameters which should be carefully tuned. Thus, better acceleration methods are still desirable and challenging.

The heavy ball (Polyak, 1964) and Nesterov's acceleration (Nesterov, 1983) are essentially an extrapolation technique. More specifically, the heavy ball is defined as

$$\mathbf{x}^{(t+1)} = (1+\tau)\mathbf{x}^{(t)} - \tau\mathbf{x}^{(t-1)} - \beta\nabla f(x^{(t)}), \tag{1}$$

where $\tau > 0$ is mixing coefficient and $\beta > 0$ is learning rate. Nesterov's acceleration is then

$$\mathbf{x}^{(t+1)} = (1+\tau)\mathbf{x}^{(t)} - \tau\mathbf{x}^{(t-1)} - \beta[(1+\tau)\nabla f(\mathbf{x}^{(t)}) - \tau\nabla f(\mathbf{x}^{(t-1)})]. \tag{2}$$

Theoretically, these extrapolation methods have been proved to be effective on (strongly) convex objective functions. Especially, because Nesterov's acceleration also uses an extrapolation gradient, it outperforms the heavy ball. However, when applied to nonconvex optimization, Nesterov's acceleration does not illustrate its ability both theoretically and empirically. In fact, Nesterov's acceleration is rarely used in optimization of DNNs.

Adam is also the state-of-the-art method for training DNNs. We note that Adam employs an accumulative sum of gradients during iterations via convex combination, which can be regarded as an interpolation gradient mechanism. This inspires us to explore a more general and more efficent interpolation scheme to accelerate gradient methods for large scale nonconvex optimization.

In this paper we consider multi-step interpolation instead of Nesterov's extrapolation. Specifically, we apply two-step and three-step interpolation schemes to optimizing DNNs such as Residual Networks with 98 layers and 200 layers. We first prespecify the mixing coefficients to conduct experiments. Although the mixing coefficients are hyperparameters to be tuned, we find that the performance of interpolation methods is insensitive to these coefficients based on the empirical results shown in Section 4. Therefore, these coefficients are actually easy to be tuned, which we consider as an advantage of interpolation methods. Addtionally, we also attempts to reduce the hyperparameters of interpolation methods. Therefore, we exploit Anderson's data-driven method (Anderson, 1965) to adaptively update the mixing coefficients on the two-step scheme during iterations.

Interestingly, experimental results show that the interpolation schemes are able to obtain acceleration, because the resulting SGD methods achieve faster convergences than the competitors such as the heavy ball and Adam methods. Moreover, our interpolation method outperforms Adam and RMSProp on the test dataset. Additionally, the performance is promoted only on ResNet-200 when the data-driven approach is used to select the mixing coefficients. Thus, the SGD methods based on interpolation are less sensitive to the mixing coefficients and are numerically stable. Therefore we feel that the interpolation scheme would be a potential acceleration approach for SGD in nonconvex optimization, especially in DNNs on large scale datasets.

The remainder of the paper is organized as follows. In Section 2 we give the details of the accelerated SGD method via interpolation. Section 3 presents an intuitive motivation of interpolation acceleration for nonconvex optimization. In Section 4 we conduct empirical analysis. Finally, we conclude our work in Section 5. The theoretical analysis of Interpolatron's convergence is shown in the appendix.

## 2. Interpolatron

We explore the multi-step interpolation scheme to accelerate gradient descent methods. Because our main focus is on training DNNs, we describe the interpolation schemes for SGD. Algorithm 1 depicts the pseudo-code of the algorithm which we refer to as Interpolatron.

The $k$-step Interpolatron updates $\mathbf{x}^{(t)}$ via the linear combination of $\mathbf{x}$ at the previous $k$ steps $\mathbf{x}^{(t-1)}, \mathbf{x}^{(t-2)}, \ldots, \mathbf{x}^{(t-k)}$ and the linear combination of gradients at the previous $k$ steps $\mathbf{g}^{(t-1)}, \mathbf{g}^{(t-2)}, \ldots, \mathbf{g}^{(t-k)}$. That is,

$$\mathbf{x}^{(t)} = \sum_{i=1}^{k} \alpha_i \mathbf{x}^{(t-i)} - \beta^{(t)} \sum_{i=1}^{k} \alpha_i \mathbf{g}^{(t-i)}. \tag{3}$$

Here $\beta$ is similar to the learning rate in the conventional gradient algorithms and should always be a positive real number. According to prior work, it is better to decay the learning rate $\beta$ (Robbins and Monro, 1951) and the decaying of $\beta$ is also common in practice. Thus we also write $\beta$ as $\beta^{(t)}$ to emphasize its value changed as $t$.

The $\alpha_i$ are the mixing coefficients satisfying $0 \leq \alpha_i \leq 1$ and $\sum_{i=1}^{k} \alpha_i = 1$. This is why we call Eqn. (3) an interpolation scheme. When $k = 2$, we can rewrite Eqn. (3) as

$$\mathbf{x}^{(t)} = (1-\alpha_2)\mathbf{x}^{(t-1)} + \alpha_2 \mathbf{x}^{(t-2)} - \beta[(1-\alpha_2)\mathbf{g}^{(t-1)} + \alpha_2 \mathbf{g}^{(t-2)}],$$

which is related to Nesterov's extrapolation in (2).

**Algorithm 1** Interpolatron
---
**Input:** Give a positive integer $k$, mixing coefficients $(\alpha_1, \alpha_2, \cdots, \alpha_k)^\top$ such as $\alpha_i \in [0,1]$ and $\sum_{i=1}^k \alpha_i = 1$, initial learning rate $\beta^{(0)}$;
Initialize $\mathbf{x}^{(j)}(-k+1 \leq j \leq 0)$ and $\mathbf{g}^{(j)}(-k+1 \leq j \leq -1)$.
**for** $t = 1$ to $T$ **do**
   Sample a mini-batch $S$ from $\{1, 2, \ldots, n\}$;
   Compute $\mathbf{g}^{(t-1)} = \frac{1}{|S|} \sum_{i \in S} \nabla f_i(\mathbf{x}^{(t-1)})$;
   Set $\mathbf{x}^{(t)} = \sum_{j=1}^k \alpha_j \mathbf{x}^{(t-j)} - \beta^{(t)} \sum_{j=1}^k \alpha_j \mathbf{g}^{(t-j)}$;
**end for**
**Output:** $\mathbf{x}^{(T)}$
---

Note that if the objective function $f(\mathbf{x})$ is $\ell$-smooth and $\mu$-strongly convex, $\tau = \frac{\sqrt{\kappa}-1}{\sqrt{\kappa}+1}$ is recommended for Nesterov's method. Because $\kappa = \frac{\ell}{\mu} > 1$ (Bubeck et al., 2015), acceleration of Nesterov's scheme for strongly convex optimization is due to extrapolation. As we have known so far, there are neither theoretical guarantees nor empirical results that Nesterov's extrapolation still works for nonconvex optimization in the literature.

Alternatively, we here study the interpolation scheme for nonconvex optimization. Considering that both the number of training samples $n$ and the dimension $d$ are very large in training DNNs, we take $k = 2$ and $k = 3$ in the following experiments. Moreover, we prespecify the values of the mixing coefficients $\alpha_i$.

However, the mixing coefficients $\alpha_i$ can be adaptively updated during iterations by Anderson's method (Walker and Ni, 2011). That is, the $\alpha_j^{(t)}$ are updated by solving the following least-squares estimation

$$\min \Big\| \sum_{i=1}^k \alpha_i^{(t)} \mathbf{g}^{(t-i)} \Big\|_2^2,$$

$$\text{s.t.} \sum_{i=1}^k \alpha_i^{(t)} = 1.$$

Let $\boldsymbol{\alpha} = (\alpha_1, \ldots, \alpha_k)^\top$, $\mathbf{1} = (1, \ldots, 1)^\top$, and $G^{(t)} = [\mathbf{g}^{(t-1)}, \mathbf{g}^{(t-2)}, \cdots, \mathbf{g}^{(t-k)}]^\top$. With some algebraic calculations, we obtain the analytic solution of $\boldsymbol{\alpha}$, which is

$$\boldsymbol{\alpha}^{(t)} = \frac{[(G^{(t)})^\top G^{(t)}]^{-1} \mathbf{1}}{\mathbf{1}^\top [(G^{(t)})^\top G^{(t)}]^{-1} \mathbf{1}}.$$

The pseudo-code of Anderson's mixing is described in Algorithm 2. Note that the resulting $\alpha_j^{(t)}$ are not necessarily all to be nonnegative. This implies that the standard Anderson's scheme inherits a hybrid of interpolation and extrapolation during iterations. It is worth noting that Anderson's scheme was originally derived for solving fixed-point problems. In this case, such a hybrid mechanism is reasonable.

To fit an interpolation scheme, we should impose the nonnegative constraint on the $\alpha_j^{(t)}$ into the above estimation problem. When $k = 2$, we have still analytic solution by projecting the above solution into $[0,1]$.

**Algorithm 2** Anderson's acceleration
---
**Input:** Give a positive integer $k$ and an initial learning rate $\beta^{(0)}$;
Initialize $\mathbf{x}^{(j)}(-k+1 \leq j \leq 0)$ and $\mathbf{g}^{(j)}(-k+1 \leq j \leq -1)$.
**for** $t = 1$ **to** $T$ **do**
   Sample a mini-batch $S$ from $\{1, 2, \ldots, n\}$;
   Compute $\mathbf{g}^{(t-1)} = \frac{1}{|S|} \sum_{i \in S} \nabla f_i(\mathbf{x}^{(t-1)})$;
   $G^{(t)} = [\mathbf{g}^{(t-1)}, \mathbf{g}^{(t-2)}, \cdots, \mathbf{g}^{(t-k)}]^\top$;
   $\mathbf{A} = [(G^{(t)})^\top G^{(t)}]^{-1}$, $1 \leq i, j \leq k$;
   $\boldsymbol{\alpha}^{(t)} = \frac{\mathbf{A}\mathbf{1}}{\mathbf{1}^\top \mathbf{A}\mathbf{1}}$
   Set $\mathbf{x}^{(t)} = \sum_{j=1}^{k} \alpha_j^{(t)} \mathbf{x}^{(t-j)} - \beta^{(t)} \sum_{j=1}^{k} \alpha_j^{(t)} \mathbf{g}^{(t-j)}$;
**end for**
**Output:** $\mathbf{x}^{(T)}$
---

## 3. Motivation and Explanation

We would show some special cases, in which extrapolation schemes do not succeed as expected, to shed light on the reason why the extrapolation-based scheme is not suitable for DNN training. Accordingly, we give an intuitive motivation to illustrate how an interpolation scheme can work well alternatively.

Some special regions on the loss surface (i.e., the surface of the loss function), such as the neighborhood of local minima and saddle points, flat regions, and cliffs (a cliff and a flat region are shown in Figure 1), are challenges for the optimizer. The gradients are approximately zero in the neighborhood of local minima, saddle points and flat regions. Another corner case is that the gradients get numerically infinite in the region of cliffs. Such phenomena are called gradient vanishing and gradient exploding. The previous study shows that such regions are common on the loss surface of neural network (Dauphin et al., 2014; Pascanu et al., 2013). The update made by SGD may be extremely small when the gradient vanishes. Conversely, an SGD update may throw the parameters very far, when the parameters reach into such a cliff region, possibly losing most of history information of the optimization (Goodfellow et al., 2016).

As the extrapolation scheme is vulnerable to such corner cases, we are motivated to propose algorithms based on interpolation to improve the optimizer's ability in cliffs (i.e., the case that gradient explodes) and flat regions (i.e., the case that gradient vanishes).

First, if the gradient vanishes, interpolation can help the parameters to escape local minima faster. For ease of exposition, we make comparison between the SGD with momentum and two-step Interpolatron on a simple function $f(\mathbf{x})$ on $\mathbb{R}$ (shown as in Figure 2(a)). The objective function $f(\mathbf{x})$ is constructed as a piecewise linear function including two sheer parts and one flat part. Despite its simplicity, it is highly related to neural networks because linear transformation layers with ReLU activation are of the similar form. The momentum is invented to improve the optimizer's ability to jump out of local minima too. The update rule of momentum is defined in Eqn. (1). The difference between two consecutive steps, $\Delta \mathbf{x}^{(t)} = \mathbf{x}^{(t)} - \mathbf{x}^{(t-1)}$, is called speed. Subtracting $\mathbf{x}^{(t)}$ on the both sides of Eqn. (1) gives

$$\begin{aligned} \Delta \mathbf{x}^{(t+1)} &= \mathbf{x}^{(t+1)} - \mathbf{x}^{(t)} \\ &= \tau(\mathbf{x}^{(t)} - \mathbf{x}^{(t-1)}) - \beta \nabla f(\mathbf{x}^{(t)}) \\ &= \tau \Delta \mathbf{x}^{(t)} - \beta \nabla f(\mathbf{x}^{(t-1)}). \end{aligned}$$

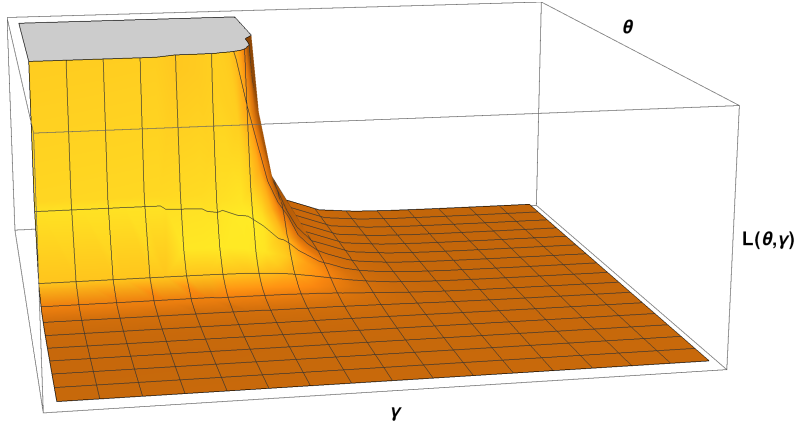

Figure 1: Cliff and flat regions in the parameter space of neural network. $\theta$ and $\gamma$ are two parameters and $L(\theta, \gamma)$ is the loss function. The region with light color is a cliff and the region with dark color is a flat region.

The term $\tau \Delta \mathbf{x}^{(t)}$ in the speed makes the previous update have impact on the next update. Thus $\alpha$ should be less than 1, otherwise the method may not converge. However, when parameters are trapped in the flat region of $f(\mathbf{x})$, a lot of steps have to be taken to jump out of the region. The problem is that although the speed term $\tau \Delta \mathbf{x}^{(t)}$ makes the parameters move forward continually when it reaches the flat region, the speed decays exponentially due to the fact that $\tau < 1$. In the specific case depicted in Figure 2, the point of parameter moves 11 steps rightward (shown as 11 blue arrows in Figure 2(b)) first. Then the point moves leftward in the next 17 steps (17 red arrows in Figure 2(b)). Finally the parameters reach the left sheer segments where the norm of the gradient is large enough. With the large gradient, the point jumps out of the flat region (the purple arrow in Figure 2(b)).

Performing much better, Interpolatron only takes 4 steps to escape the local minima and the flat region (shown as 4 arrows in Figure 2(c)). The better performance of Interpolatron might be attributed to the interpolation of gradients. When the point of parameter reaches the flat region after the first step, the interpolation of gradients, (i.e. $\alpha_1 \mathbf{g}^{(t-1)} + \alpha_2 \mathbf{g}^{(t-2)}$), makes the large gradient on sheer region effective. Thus the point still moves a large step to the right side.

Second, in the case of gradient exploding, the interpolation of $\mathbf{x}^{(t-1)}$ and $\mathbf{x}^{(t-2)}$ may avoid catapulting the parameters to a point very far away, which makes the previously done optimization work lost. Any algorithm that introduces extrapolation, such as the momentum, performs even worse, because one giant update step would have persistent impact on several

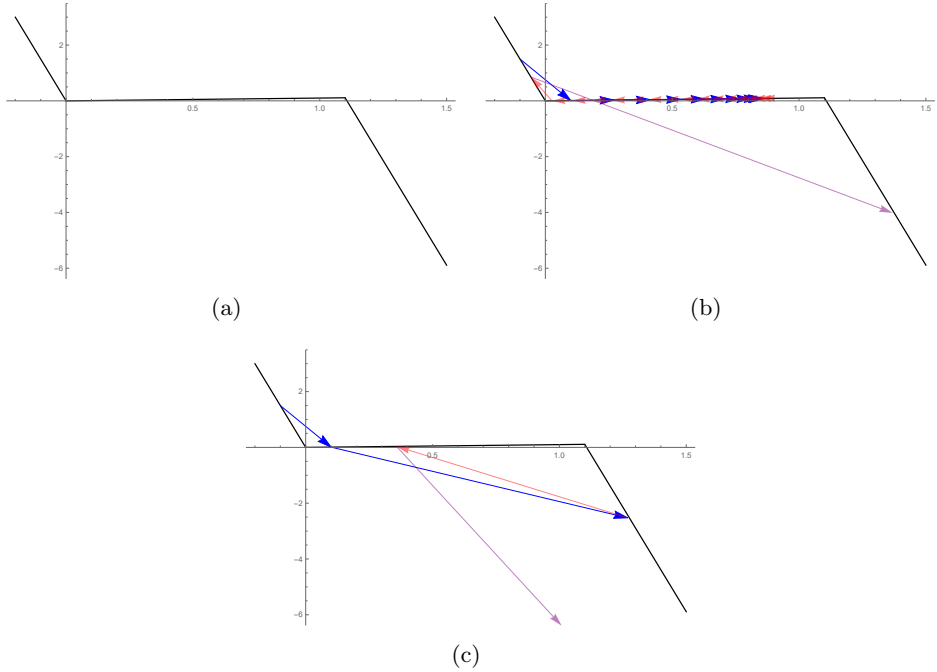

Figure 2: (a) The function $f(x) \in \mathbb{R}$. It is a piecewise linear function. The slope of the left and the right sheer segments is -15 and the slope of the middle flat segment is 0.1. (b) The trajectory of SGD with momentum. It takes 29 steps to escape the local minima. The steps that move to the right are marked as blue arrows and the steps that move to the left are marked as red arrows. The final step which leads to escaping the local minima is marked as purple arrow. (c) The trajectory of Interpolatron. The color scheme is the same as in (b). It takes 4 steps to escape the local minima. The endpoint of the last step is too far and thus the arrow is truncated for making the figure size at proper level. The last step moves to $x = 2.527$, where $f(x) = -21.393$.

consequent updates. Observing the update rule of Interpolatron, we obtain

$$\Delta \mathbf{x}^{(t)} = -\alpha_2 \Delta \mathbf{x}^{(t-1)} - \beta(\alpha_1 \mathbf{g}^{(t-1)} + \alpha_2 \mathbf{g}^{(t-2)}).$$

The operation of interpolation would cancel the giant step partly.

Similarly, we demonstrate the advantage of interpolation in a simple case, where $f(\mathbf{x})$ on $\mathbb{R}$ is shown as in Figure 3(a). At this time, our concern is about whether the parameters will jump too far, making the previous optimization work futile. After two steps, Interpolatron still stays in the region near to the minima while SGD with momentum gets to $\mathbf{x} = 14.806$, a point very far. Four steps later, Interpolatron goes backward while SGD with momentum goes even further. In fact, SGD with momentum moves to $\mathbf{x} = 50.441$ after 8 steps and never goes back to the initial position again.

Although it seems that the two problems could be avoided if the learning rate for SGD with momentum is set up properly, the key point is that setting up the proper learning rate is hard or even impossible. In most situations, a larger learning rate is required to escape local minima faster, while the larger learning rate will make the performance of SGD with momentum even worse when the gradient explodes. Similarly, a smaller learning rate would alleviate the gradient exploding but makes escaping local minima slower. As a result, there exists tradeoff between increasing and decreasing the learning rate. The so-called proper learning rate for SGD with momentum may even not exist.

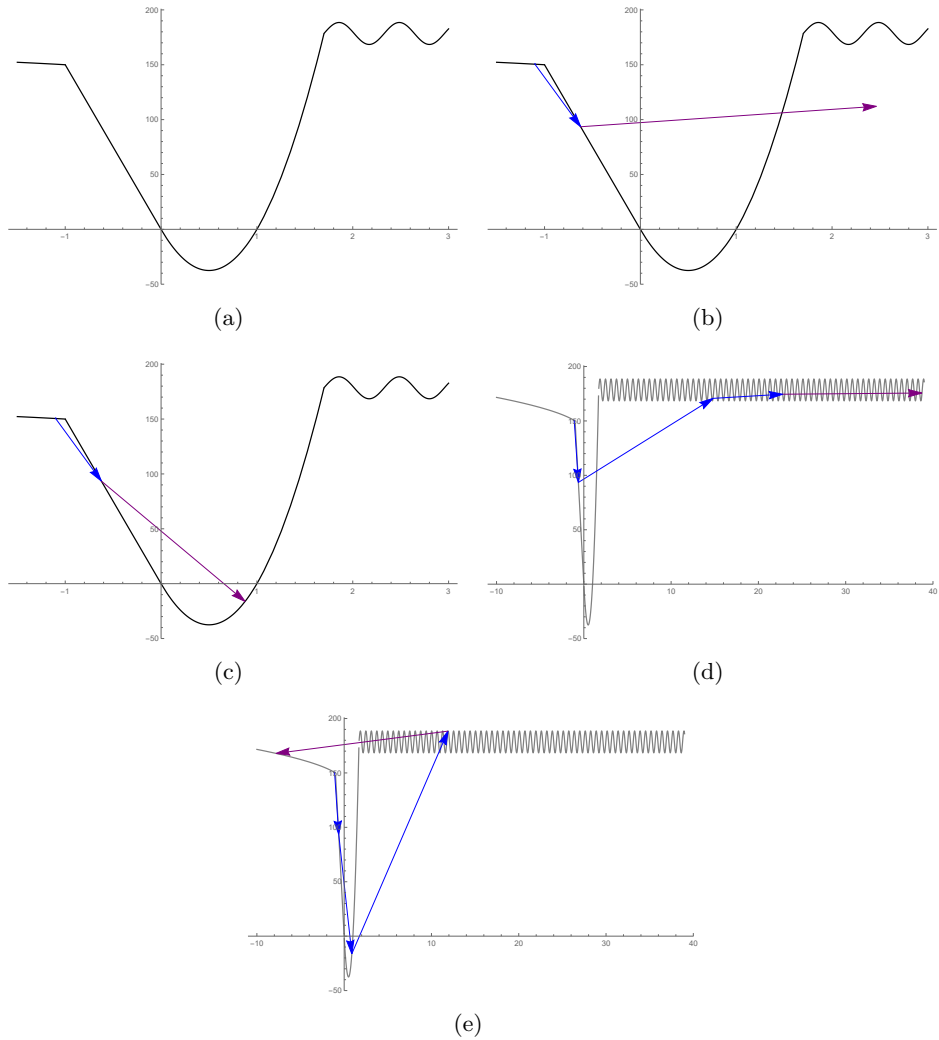

Figure 3: (a) The function $f(\mathbf{x}) : \mathbb{R} \to \mathbb{R}$. The first step is initialized as a step of SGD for both the algorithms. (b) The trajectory of SGD with momentum in 2 steps. The final step marked as purple arrow is truncated to make the figure size at proper level. The last step moves to $\mathbf{x} = (14.806)$. (c) The trajectory of Interpolatron in 2 steps. The color scheme is the same as in (b). (d) The trajectory of SGD with momentum in 4 steps. (e) The trajectory of Interpolatron in 4 steps.

Table 1: Hyperparameters for ResNet-98 on CIFAR-10

| Optimizer | learning rate($\beta^{(0)}$) | $\alpha$ ($\tau$) |
|---|---|---|
| SGD | 0.25 | N/A |
| Adam | $5.0 \times 10^{-4}$ | N/A |
| Momentum | 0.025 | $\tau = 0.9$ |
| Nesterov | 0.025 | $\tau = 0.9$ |
| 2-nd Interpolatron | 0.1 | $(0.05, 0.95)$ |
| 3-rd Interpolatron | 0.1 | $(0.1, 0.3, 0.6)$ |
| Anderson | 0.25 | N/A |

## 4. Experiments

In this section we validate the performance of Interpolatron on DNNs and large scale data sets. Residual networks (ResNet) (He et al., 2016) have been proved successful in many areas and are a great challenge to optimizer due the extremely deep architecture. We thus choose ResNet with layers of various number as the training model.

We implement the two-step and three-step Interpolatrons as well as Anderson's acceleration (two-step). We also conduct comparison with SGD, SGD with momentum (SGD-m)(Polyak, 1964), Adam, and Nesterov's accelerated gradient (Nesterov, 1983). The parameters $\mathbf{x}^{(j)}(-k+1 \leq j \leq -1)$ and gradients $\mathbf{g}^{(j)}(-k+1 \leq j \leq -1)$ are initialized randomly. Moreover, $\mathbf{x}^{(0)}$ is initialized by variance scaling, which is the same as the ways other optimizers initialize. Based on the results of experiments, truncated normal distribution is recommended. The code will be available.

### 4.1 Residual Network on CIFAR-10

The CIFAR-10 dataset includes 60,000 color images of size $32 \times 32$ in 10 classes. We use 50,000 images for training and the rest 10,000 for test.

We train 98-layer and 200-layer ResNets on CIFAR-10. The learning rates of SGD, SGD-m, Nesterov's accelerated gradient, and Adam are carefully tuned separately. The hyperparameters are summarized in Tables 1 and 2. The objective function consists of the cross entropy and the regularization of weight decay (Krogh and Hertz, 1992); that is,

$$f(\mathbf{x}) \triangleq \sum_{i=1}^{n} f_i(\mathbf{x}) + \lambda \|\mathbf{x}\|^2,$$

where $\lambda$ is set as $2 \times 10^{-4}$ in our experiments. The size of mini batch $S$ is 128. The total number of epochs is 250. The $\beta$ is divided by ten after 100, 150, and 200 epochs. We do not use dropout. All the methods in this section are trained on a single GTX 1080 Ti GPU.

The loss curves and accuracy curves on the training set are shown in Figure 4. On the 98-layer ResNet, three-step Interpolatron achieves the most impressive acceleration. Two-step Interpolatron and two-step Anderson's acceleration also illustrate better ability than the others. On the 200-layer ResNet, two-step Interpolatron, two-step Anderson's acceleration and three-step Interpolatron are still the best ones and their advantage over the rest ones is even significant. For the three interpolation schemes, the accuracy lies in the acceptable range.

Table 2: Hyperparameters for ResNet-200 on CIFAR-10

| Optimizer | learning rate($\beta^{(0)}$) | $\alpha$ ($\tau$) |
|---|---|---|
| SGD | 0.25 | N/A |
| Adam | $1.0 \times 10^{-3}$ | N/A |
| Momentum | 0.05 | $\tau = 0.9$ |
| Nesterov | 0.05 | $\tau = 0.9$ |
| 2-nd Interpolatron | 0.25 | $(0.1, 0.9)$ |
| 3-rd Interpolatron | 0.25 | $(0.1, 0.3, 0.6)$ |
| Anderson | 0.1 | N/A |

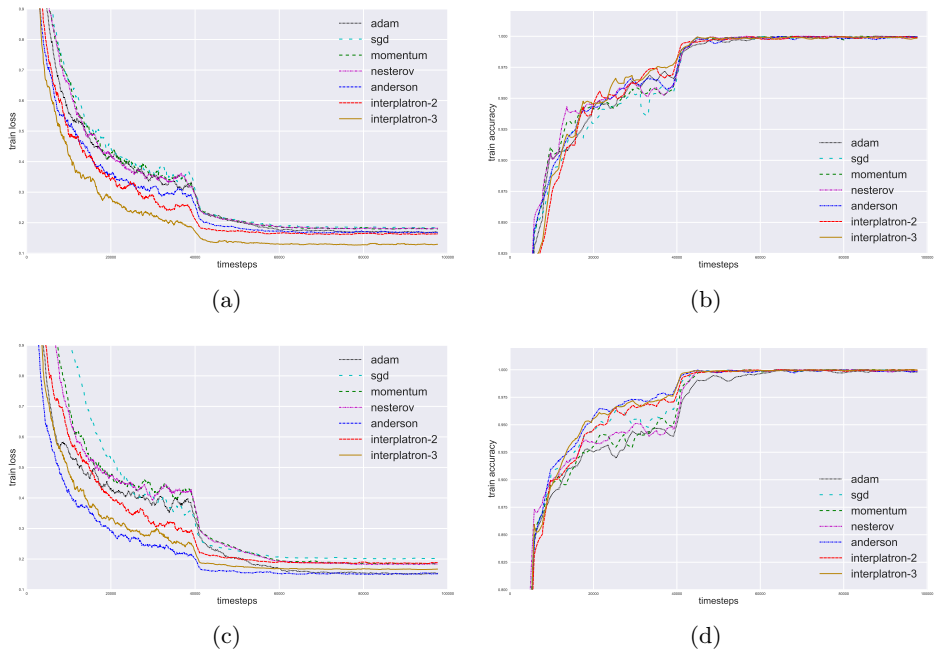

Figure 4: The results of ResNet-98 and ResNet-200 on CIFAR-10. (a) The loss curves of ResNet-98 on training set. (b) The accuracy curves of ResNet-98. (c) The loss curves of ResNet-200 on training set. (d) The accuracy curves of ResNet-200.

Additionally, we would point out that during the training of the conventional Anderson's accelerated gradient, the obtained $\alpha_1^{(t)}$ and $\alpha_2^{(t)}$ take the values in $[0, 1]$ in most cases (although the $\alpha_j^{(t)}$ are not constrained in interval $[0, 1]$). Thus in most of iterations, Anderson's acceleration also works in an interpolation scheme. This implies that interpolation is more suitable for training neural networks. When the scale of the network is large enough (e.g., 200-layer ResNet), Anderson's acceleration performs the best.

Table 3: Hyperparameters for ResNet-50 on ImageNet

| Optimizer | learning rate($\beta^{(0)}$) | $\alpha$ ($\tau$) |
|---|---|---|
| SGD | 0.2 | N/A |
| Adam | $1.0 \times 10^{-3}$ | N/A |
| Momentum | 0.2 | $\tau = 0.9$ |
| 2-step Interpolatron | 0.1 | $(0.1, 0.9)$ |
| 3-step Interpolatron | 0.1 | $(0.1, 0.3, 0.6)$ |
| Anderson | 0.1 | N/A |

### 4.2 Residual Network on ImageNet

ImageNet (Russakovsky et al., 2015) is one of the most large datasets currently. There are more than one million images of 1,000 classes in ImageNet. We train the models with 1, 281, 167 images and use the rest for test. To be more exact, by mentioning ImageNet, we actually mean the classification subtask of ImageNet Large Scale Visual Recognition Challenge (ILSVRC) 2012 here.

The architecture of the networks selected is 50-layer ResNet. The hyperparameters of the methods are described in Table 3. The coefficient for weight decay is taken as $10^{-4}$. The batch size is 512. We iterate 90 epochs in total and $\beta$ is divided by 10 every 30 epochs. All the methods in this section are trained on 8 GTX 1080 Ti GPUs. The loss curves, top-1 and top-5 accuracy curves are shown in Figure 5. Two-step Interpolatron and three-step Interpolatron demonstrate acceleration as expected, especially in the early stage. The difference between full gradient on the whole data and stochastic gradient computed on a mini-batch may be very large when dataset is ultimately large. Because the mixing coefficients of Anderson's acceleration are obtained by least-squares estimation, its performance may be damaged seriously if the stochastic gradient diverges from the full gradient a lot. The top-1 and top-5 accuracies with the two-step Interpolatron and three-step Interpolatron are also good among all the optimizers tested.

### 4.3 Sensitivity to Hyperparameters

In tuning the hyperparameters, we find that the performance of Interpolatron is insensitive to the values of the hyperparameters, which we consider as an advantage. To justify our assertion, we conduct experiments exploring the impacts of the different hyperparameters on Interpolatron's performance. We train 98-layer ResNet on CIFAR-10 with different configuration of hyperparameters. The setting of weight decay and learning rate decay are the same with the settings in Section 4.1. First, we fix $\beta^{(0)}$ (denoted $\beta$ for simplicity), test its performance on various $\alpha$ ($\alpha_1$ is selected from $0.05, 0.1, 0.25, 0.5$). Figures 6(a), 6(b), and 6(c) depict that the different values of $\alpha$ have relatively little impact on the performance. Then, we fix $\alpha$ and test its performance on various $\beta^{(0)}$ (also denoted $\beta$). Now $\beta$ is selected from $0.05, 0.1, 0.25$. Figures 7(a), 7(b), and 7(c) show that Interpolatron is not sensitive to the value of $\beta$ too. But Figure 7(d) shows that SGD relies heavily on the learning rate.

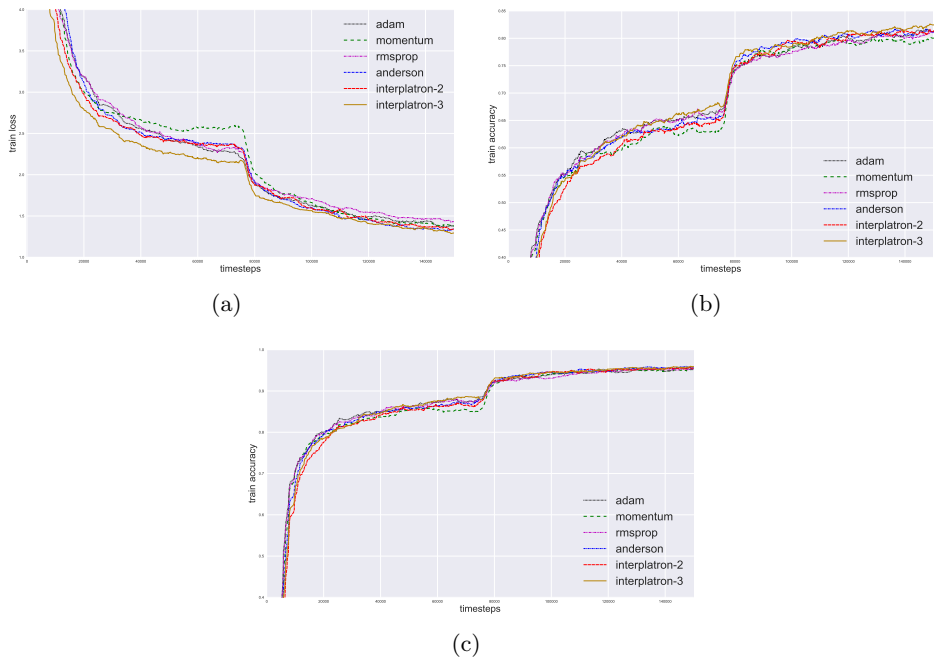

Figure 5: The results of ResNet-50 on ImageNet. (a) The loss curves of ResNet-50. (b) The top-1 accuracy curves of ResNet-50. (c) The top-5 accuracy curves of ResNet-50.

## 5. Concluding Remarks

In this paper we have introduced interpolation acceleration schemes into SGD methods for training a deep neural network, which corresponds to a large-scale and high-dimensional nonconvex optimization problem. In particular, we have proposed the interpolation-based SGD method and called it Interpolatron. We have conducted the empirical analysis and comparison with the existing benchmark methods, on DNNs and large datasets. The experimental results have shown that Interpolatron is able to obtain acceleration. Compared with the SGD with momentum and Adam, Interpolatron converges much faster and does not bring extra computations. Moreover, it is not sensitive to the choice of the mixing coefficients. In general, the mixing coefficients are just prespecified by users, although they can be adaptively updated based on Anderson's method.

We have proved the convergence of the interpolation scheme under the assumptions of smoothness and strongly convexity theoretically (the detail is given in the supplementary materials). However, for the nonconvex case, the convergence is still open. Recently, Scieur et al. (2017) discussed the relationship between integration methods and optimization algorithms, shedding light on the connection between acceleration of optimization and numerical ordinary differential equation. This connection would provide us an potential approach for the convergence analysis of the interpolation scheme in nonconvex optimization. We will dig out this issue in future. In addition to the Convolutional Neural Network, we will also apply the interpolation schemes to Recurrent Neural Network to illustrate their ability.

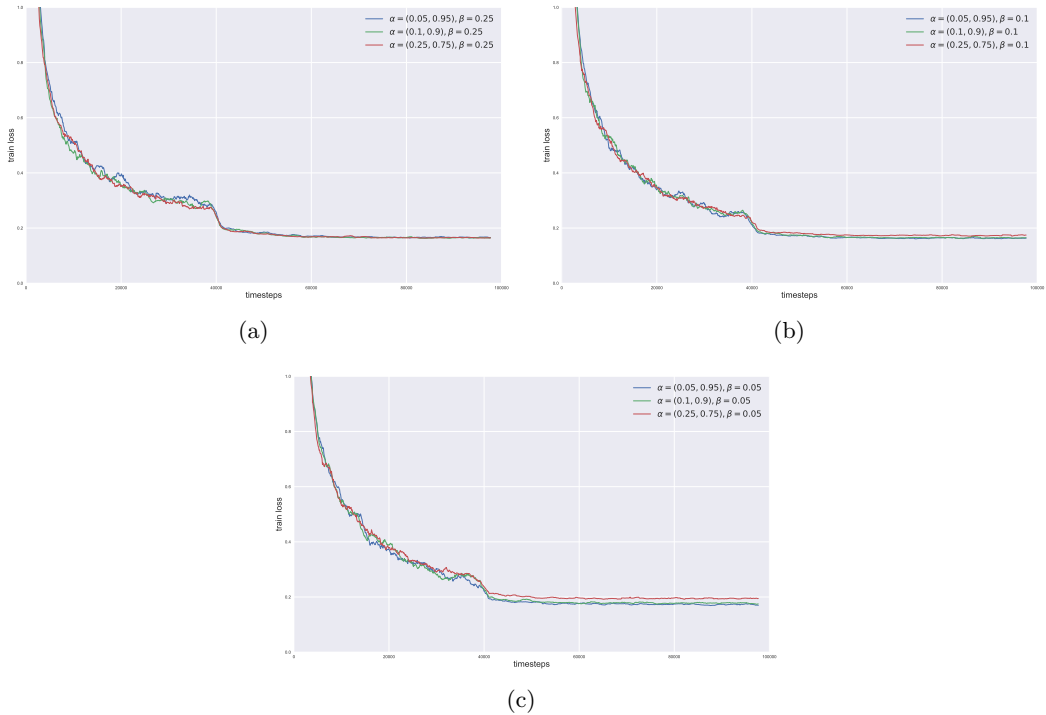

Figure 6: The architecture of neural network is ResNet-98 and dataset is CIFAR-10. (a) The loss curve of Interpolatron when fix $\beta = 0.25$ and select $\alpha_1$ from 0.05, 0.1, 0.25. (b) The loss curve of Interpolatron when fix $\beta = 0.1$ and select $\alpha_1$ from 0.05, 0.1, 0.25. (c) The loss curve of Interpolatron when fix $\beta = 0.05$ and select $\alpha_1$ from 0.05, 0.1, 0.25.

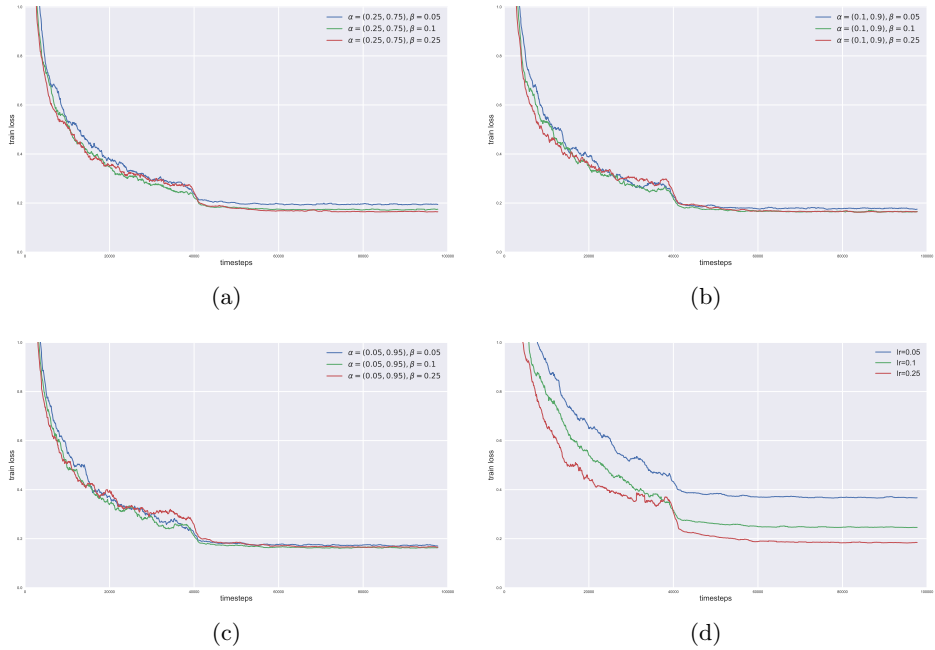

Figure 7: The architecture of neural network is Resnet-98 and dataset is CIFAR-10. (a) The loss curve of Interpolatron when fix $\alpha = (0.25, 0.75)$ and select $\beta$ from $0.05, 0.1, 0.25$. (b) The loss curve of Interpolatron when fix $\alpha = (0.1, 0.9)$ and select $\beta$ from $0.05, 0.1, 0.25$. (c) The loss curve of Interpolatron when fix $\alpha = (0.05, 0.95)$ and select $\beta$ from $0.05, 0.1, 0.25$. (d) The loss curve of SGD when select $\beta$ from $0.05, 0.1, 0.25$.

## Appendix A. Proof of Convergence

In this section, we will show the interpolation-based gradient descent's convergence in the smooth and stronly convex case.

We assume that $f : \mathbb{R}^d \to \mathbb{R}$ is

- twice continuously differentiable
- smooth with constant $\eta > 0$:

$$\|\nabla f(x) - \nabla f(y)\|_2 \leq \eta \|x - y\|_2 \tag{4}$$

  or equivalently

$$\nabla^2 f(x) \preceq \eta I, \forall x \tag{5}$$

- strongly convex with constant $\mu > 0$:

$$f(y) \geq f(x) + \nabla f(x)^T (y - x) + \frac{\mu}{2} \|y - x\|_2^2 \tag{6}$$

  or equivalently

$$\nabla^2 f(x) \succeq \mu I, \forall x \tag{7}$$

**Lemma 1** *for any $x, y$, there exists a $H$ which satisfies $\mu I \preceq H \preceq \eta I$ and*

$$\nabla f(x) - \nabla f(y) = H(x - y)$$

**Proof**

$$\nabla f(x) - \nabla f(y) = \left( \int_0^1 \nabla^2 f(y + t(x - y)) \mathrm{d}t \right) (x - y)$$

Let

$$H = \int_0^1 \nabla^2 f(y + t(x - y)) \mathrm{d}t,$$

and notice that for any $u \in \mathbb{R}^d$,

$$u^\top H u = \int_0^1 u^\top \nabla^2 f(y + t(x - y)) u \, \mathrm{d}t$$
$$\leq \int_0^1 \eta \|u\|_2^2 \mathrm{d}t$$
$$\leq \eta \|u\|_2^2.$$

So we can get $H \preceq \eta I$, and $H \succeq \mu I$ holds for the similar reason. ∎

Then given $x_1, x_2, \cdots, x_m$, define $x_{k+m}$ as

$$x_{k+m} = \sum_{i=1}^{m} \alpha_i x_{k+m-i} - \sum_{i=1}^{m} \alpha_i \nabla f(x_{k+m-i})$$

for $k > 0$. Denote the optimal point of $f$ by $x^*$. First, following Lemma 1, we get

$$\begin{bmatrix} x_{k+m} - x^* \\ x_{k+m-1} - x^* \\ \vdots \\ x_{k+1} - x^* \end{bmatrix}$$

$$= \begin{bmatrix} \sum_{i=1}^{m} \alpha_i(x_{k+m-i} - x^*) \\ x_{k+m-1} - x^* \\ \vdots \\ x_{k+1} - x^* \end{bmatrix} - \beta \begin{bmatrix} \sum_{i=1}^{m} \alpha_i \nabla f(x_{k+m-i}) \\ 0 \\ \vdots \\ 0 \end{bmatrix}$$

$$= \begin{bmatrix} \alpha_1 I & \alpha_2 I & \cdots & \alpha_{m-1} I & \alpha_m I \\ I & 0 & \cdots & 0 & 0 \\ \vdots & \vdots & & \vdots & \vdots \\ 0 & 0 & \cdots & I & 0 \end{bmatrix} \begin{bmatrix} x_{k+m-1} - x^* \\ x_{k+m-2} - x^* \\ \vdots \\ x_k - x^* \end{bmatrix} - \beta \begin{bmatrix} \sum_{i=1}^{m} \alpha_i H_{k+m-i}(x_{k+m-i} - x^*) \\ 0 \\ \vdots \\ 0 \end{bmatrix}$$

$$= \begin{bmatrix} \alpha_1 J_1 & \alpha_2 J_2 & \cdots & \alpha_{m-1} J_{m-1} & \alpha_m J_m \\ I & 0 & \cdots & 0 & 0 \\ \vdots & \vdots & & \vdots & \vdots \\ 0 & 0 & \cdots & I & 0 \end{bmatrix} \begin{bmatrix} x_{k+m-1} - x^* \\ x_{k+m-2} - x^* \\ \vdots \\ x_k - x^* \end{bmatrix},$$

where $J_i = I - \beta H_{k+m-i}$, $\mu I \preceq H_{k+m-i} \preceq \eta I$. For brief, denote

$$A_k = \begin{bmatrix} \alpha_1 J_1 & \alpha_2 J_2 & \cdots & \alpha_{m-1} J_{m-1} & \alpha_m J_m \\ I & 0 & \cdots & 0 & 0 \\ \vdots & \vdots & & \vdots & \vdots \\ 0 & 0 & \cdots & I & 0 \end{bmatrix}. \tag{8}$$

With matrix calculation, we get

$$\det |A_k - \lambda I_{m \times d}| = \pm \det \left| (\sum_{i=1}^{m} \lambda^{m-i} \alpha_i J_i) - \lambda^m I \right|, \tag{9}$$

where $\lambda \in \mathbb{C}$.

**Lemma 2** *Given $\rho_i \in \mathbb{R}, (i = 1, \cdots, m), |\rho| > \sum_{i=1}^{m} |\rho_i|$, there exists $0 < \xi < 1$, such that $\forall z \in \mathbb{C}, |z| > \xi$,*

$$|\rho||z|^m > \sum_{i=1}^{m} |\rho_i||z^{m-i}|$$

*holds.*

**Proof** Let
$$p(x) = |\rho|x^m - (\sum_{i=1}^{m} |\rho_i|x^{m-i}), x \in \mathbb{R}.$$

In fact, for $x \geq 1$,
$$|\rho|x^m > \sum_{i=1}^{m} |\rho_i|x^m \geq \sum_{i=1}^{m} |\rho_i|x^{m-i},$$

i.e. $p(x) > 0$. Combined with $p(0) = 0$, the largest real root of $p(x)$, denoted by $\zeta$, lies in $[0, 1)$. Let $\xi = \frac{\zeta+1}{2}$, and the conclusion holds. ∎

**Lemma 3** *Supposing that $\theta = \max\{|1 - \beta\mu|, |1 - \beta\eta|\} < 1$, there exists $0 < \xi < 1$, such that the spectral radius of $A_k$ is less than $\xi$.*

**Proof** According to Eqn. (9), the conclusion to be proved is equvalent to
$$\det\left|(\sum_{i=1}^{m} \lambda^{m-i}\alpha_i J_i) - \lambda^m I\right| \neq 0, \forall |\lambda| \geq \xi.$$

Note that for $\forall y \in \mathbb{C}^d$, we have
$$\bar{y}^\top(\sum_{i=1}^{m} \lambda^{m-i}\alpha_i J_i - \lambda^m I)y = (\sum_{i=1}^{m} \lambda^{m-i}\alpha_i \bar{y}^\top y) - \lambda^m \bar{y}^\top y - \beta(\sum_{i=1}^{m} \lambda^{m-i}\alpha_i \bar{y}^\top H_{k+m-i} y).$$

Now let
$$\rho_i = \alpha_i(\bar{y}^\top y - \beta\bar{y}^\top H_{k+m-i} y).$$

According to $\mu I \preceq H_{k+m-i} \preceq \eta I$, we have
$$\mu\bar{y}^\top y \leq \bar{y}^\top H_{k+m-i} y \leq \eta\bar{y}^\top y.$$

Thus $|\rho_i| \leq \alpha_i \theta \bar{y}^\top y$ and $\sum_{i=1}^{m} |\rho_i| \leq \theta \bar{y}^\top y < \bar{y}^\top y$ hold.

Applying lemma (2), there exists $\xi, 0 < \xi < 1$, such that for $|\lambda| > \xi$
$$|\lambda^m| > \sum_{i=1}^{m} \alpha_i \theta |\lambda^{m-i}| \geq (\sum_{i=1}^{m} |\rho_i||\lambda^{m-i}|)/\bar{y}^\top y,$$

so
$$\bar{y}^\top(\sum_{i=1}^{m} \lambda^{m-i}\alpha_i J_i - \lambda^m I)y \neq 0.$$

Due to the arbitrariness of $y$, $(\sum_{i=1}^{m} \lambda^{m-i}\alpha_i J_i) - \lambda^m I$ is nonsingular, Q.E.D. ∎

**Theorem 4** *There exists $\xi, 0 < \xi < 1, D_0 > 0$, such that*
$$\|x_{k+m} - x^*\| \leq \xi^k D_0 \tag{10}$$

**Proof** According to Lemma (3) and the property of spectral radius, there exists $\xi'$, $0 < \xi' < 1$ and a matrix norm $\|\cdot\|_*$, such that $\|A_k\|_* \leq \frac{1+\xi'}{2} = \xi$. Moreover, by the equivalence of the square matrix norm, there exists a constant $C$, such that $\|M\| \leq C\|M\|_*$. Thus,

$$\|A_1 A_2 \cdots A_k\| \leq C\|A_1 A_2 \cdots A_k\|_* \leq C\|A_1\|_* \|A_2\|_* \cdots \|A_k\|_* \leq C\xi^k,$$

and

$$\left\| \begin{bmatrix} x_{k+m} - x^* \\ x_{k+m-1} - x^* \\ \vdots \\ x_{k+1} - x^* \end{bmatrix} \right\| \leq \|A_1 A_2 \cdots A_k\| \left\| \begin{bmatrix} x_m - x^* \\ x_{m-1} - x^* \\ \vdots \\ x_1 - x^* \end{bmatrix} \right\| \leq C\xi^k \left\| \begin{bmatrix} x_m - x^* \\ x_{m-1} - x^* \\ \vdots \\ x_1 - x^* \end{bmatrix} \right\|.$$

Then we get the convergence of $x$. ∎